\documentclass[conference]{IEEEtran}
\IEEEoverridecommandlockouts
\usepackage{cite}
\usepackage{amsmath,amssymb,amsfonts}
\usepackage[linesnumbered,ruled,vlined]{algorithm2e}
\usepackage{algorithmic}
\usepackage{graphicx}
\usepackage{textcomp}
\usepackage{xcolor}
\usepackage{bm}
\usepackage[font=footnotesize]{caption}
\usepackage{dsfont}
\usepackage[font=footnotesize]{subcaption}
\usepackage{url}
\usepackage{amsthm}

\newcommand{\norm}[1]{\left\lVert#1\right\rVert}

\newcommand{\tr}{\mathsf{tr}}

\newcommand{\Diag}{\mathsf{Diag}}

\newcommand{\bS}{\bm{S}}

\newcommand{\bX}{\bm{X}}

\def\BibTeX{{\rm B\kern-.05em{\sc i\kern-.025em b}\kern-.08em
    T\kern-.1667em\lower.7ex\hbox{E}\kern-.125emX}}
\begin{document}

\title{Learning Undirected Graphs in Financial Markets\\
\thanks{This work was supported by the Hong Kong GRF 16207019 research grant.}
}

\author{\IEEEauthorblockN{Jos\'e Vin\'icius de Miranda Cardoso}
\IEEEauthorblockA{\textit{Department of Electronic and Computer Engineering} \\
\textit{The Hong Kong University of Science and Technology}\\
Clear Water Bay, Hong Kong \\
$\mathsf{jvdmc@connect.ust.hk}$}
\and
\IEEEauthorblockN{Daniel P. Palomar}
\IEEEauthorblockA{\textit{Department of Electronic and Computer Engineering} \\
\textit{Department of Industrial Engineering and Decision Analytics}\\
\textit{The Hong Kong University of Science and Technology}\\
Clear Water Bay, Hong Kong\\
$\mathsf{palomar@ust.hk}$}
}

\maketitle

\begin{abstract}
We investigate the problem of learning undirected graphical models under Laplacian structural
constraints from the point of view of financial market data. We show that Laplacian constraints
have meaningful physical interpretations related to the market index factor and to conditional
correlations between stocks. Those interpretations lead to a set of guidelines that users should
be aware of when estimating graphs in financial markets. In addition, we propose algorithms to
learn undirected graphs that account for stylized facts and tasks intrinsic to financial data such as
non-stationarity and stock clustering.
\end{abstract}


\section{Introduction}

Learning the structure of general graphical models
is an NP-hard task~\cite{anima2012} whose importance is critical towards visualizing, understanding, and leveraging
the full potential contained in the data that live in such structures. Learning graphs from data is a fundamental problem
in the statistical graph learning and signal processing
fields~\cite{lake2010, egilmez2017, zhao2019}, %witten2011, pavez2018,
having a direct impact on applications such as unsupervised learning,
clustering, and %hao2018}%, kumar20192}, % hsieh2012, sun14, tan2015, 
applied finance~\cite{feiping2016,  kumar2019, marti2017}. %mantegna1999, deprado2016, network topology inference~\cite{segarra2017, mateos2019, geert2019},
%and graph neural nets~\cite{wu2019}.

Nonetheless, most existing techniques for learning graphs are often unable to impose a particular
graph structure due to their inability to incorporate prior information in the
learning process.
Moreover, most graph learning frameworks are designed
towards static %(non time-varying)
networks~\cite{dong2016, kalofolias2016, egilmez2017, zhao2019} % structures,
which inherently neglect dynamic time-domain variations in real %network
data. As a consequence,
they usually lack practicality especially in non-stationary data regimes, which is often the case
in data from financial stock markets.

Motivated by practical applications such as clustering of stocks and understanding their time-domain
variations, we investigate the problem of estimating graph matrices whose structure follow those of Laplacian
matrices of undirected weighted graphs in a financial context both for static and dynamic graphs.

The main contributions of our paper are as follows:
\begin{enumerate}
  \item We for the first time provide natural interpretations for the Laplacian constraints of graphs estimated from stock market data.
  This leads to meaningful and intuitive guidelines on the data processing required prior to learning graphs. 
  \item We show that rank constraints alone, a practice often used by state-of-the-art methods, are not sufficient to learn $k$-component graphs.
  \item We propose novel formulations to learn: i) $k$-component graphs and ii) time-varying graphs.
  \item We develop simple trading strategies as a result of the estimated time-varying graphs.
\end{enumerate}

\section{Background and Related Work}

A graph is denoted as a triple $\mathcal{G} = \left(\mathcal{V}, \mathcal{E}, \bm{W}\right)$,
where $\mathcal{V} = \left\{1, 2, \dots, p\right\}$ is the vertex (or node) set,
$\mathcal{E} \displaystyle \subseteq \left\{\left\{u, v\right\}: u,v \in \mathcal{V}\right\}$ is the edge set, that
is, a subset of the set of all possible unordered pairs of $p$ nodes
such that $\{u, v\} \in \mathcal{E}$ if and only if nodes $u$ and $v$ are connected.
We denote the number of elements in $\mathcal{E}$ by $\vert\mathcal{E}\vert$.
$\bm{W} \in \mathbb{R}_{+}^{p\times p}$ is the symmetric weighted adjacency matrix that satisfies
$W_{ii} = 0, W_{ij} > 0 ~\text{if and only if}~ \{i,j\} \in \mathcal{E} ~\text{and} ~W_{ij} = 0~\text{otherwise}$.
The graph Laplacian matrix $\bm{L}$ is defined as $\bm{L} \triangleq \bm{D} - \bm{W}$, where
$\bm{D} \triangleq \Diag(\bm{W}\mathbf{1})$ is the degree matrix.

An attractive improper Gaussian Markov Random Field (GMRF)~\cite{rue2005} %slawski2014,
is denoted as a $p$-dimensional, real-valued, Gaussian
random variable $\bm{x}$ with mean vector $\boldsymbol{\mu}$ and low-rank precision (inverse covariance) matrix $\boldsymbol{\Xi}$.
The data generating process is assumed to be a zero-mean, attractive improper GMRF $\bm{x} \in \mathbb{R}^{p}$, 
such that $x_i$ is the random variable generating a signal measured at node $i$,
whose low-rank precision matrix is modeled as a graph Laplacian matrix.
Assume we are given $n$ observations from $\bm{x}$, \textit{i.e.}, $\bm{X} \in \mathbb{R}^{n\times p}$,
$\bm{X} = \left[\bm{x}_{1}, \bm{x}_{2}, \dots, \bm{x}_p\right]$, $\bm{x}_{i} \in \mathbb{R}^{n}$.
The Penalized Maximum Likelihood Estimator of the precision matrix of $\bm{x}$, on the basis of the observed data $\bm{X}$, is
\begin{equation}
\begin{array}{ll}
  \underset{\bm{L} \succeq \mathbf{0}}{\mathsf{minimize}} & \mathsf{tr}\left(\bm{L}\bm{S}\right)
  - \mathsf{log~gdet}\left(\bm{L}\right) + h_{\boldsymbol\alpha}(\bm{L}),\\
  \mathsf{subject~to} & \bm{L}\mathbf{1} = \mathbf{0},~L_{ij} = L_{ji} \leq 0,
\end{array}
\label{eq:laplacian-learning}
\end{equation}
where $\bm{S}$ is a similarity matrix, such as sample covariance (or correlation) matrix $\bm{S} \propto \bm{X}^\top\bm{X}$, $\mathsf{gdet}(\bm{L})$
is the pseudo determinant of $\bm{L}$, \textit{i.e.}, the product of its positive eigenvalues~\cite{knill2014},
and $h_{\boldsymbol\alpha}(\bm{L})$ is a regularization function, with hyperparameter vector $\boldsymbol\alpha$,
to promote certain properties on $\bm{L}$ such as sparsity.

Problem~\eqref{eq:laplacian-learning} is a fundamental problem in the graph signal processing field
that has served as a cornerstone for many extensions, primarily those involving the inclusion
of structure onto $\bm{L}$~\cite{egilmez2017,kumar2019}. %,kumar20192}%,kumar20193}. %,pavez2018,
Even though Problem~\eqref{eq:laplacian-learning} is convex, assuming $h_{\boldsymbol\alpha}(\cdot)$ is convex, it is not
adequate to be solved by disciplined convex programming languages, such as $\mathsf{cvxpy}$~\cite{diamond2016cvxpy}, due to
scalability issues related to the computation of $\mathsf{log~gdet}(\bm{L})$. Indeed, recently a few works
have proposed algorithms based on Majorization-Minimization (MM)~\cite{sun2017}, and Alternating Direction Method of Multipliers
(ADMM)~\cite{boyd2011} to solve Problem~\eqref{eq:laplacian-learning} in an efficient and scalable fashion~\cite{egilmez2017, zhao2019}.

On the other hand, due to nuisances involved in dealing with the term $\mathsf{log ~gdet}(\bm{L})$,
several works focused on the assumption that the underlying signals in a graph are smooth~\cite{kalofolias2016, dong2016}. %,dong2016,chepuri2017}.
In its simplest form, learning a smooth graph from a data matrix
$\bm{X} \in \mathbb{R}^{n \times p}$
is tantamount to finding an adjacency matrix $\bm{W}$ that minimizes the
Dirichlet energy:
\begin{equation}
  \begin{array}{ll}
   \underset{\bm{W}}{\mathsf{minimize}} &
  \frac{1}{2}\sum_{i,j}W_{ij}\norm{\bm{x}_i - \bm{x}_j}^2_2 \\
  \mathsf{subject~to} & W_{ij} = W_{ji} \geq 0, \mathsf{diag}(\bm{W}) = \mathbf{0}.
  \end{array}
  \label{eq:signal-smoothness1}
\end{equation}

In order for Problem~\eqref{eq:signal-smoothness1} to be well-defined,
several constraints have been proposed in the literature.
For instance, Kalofolias~\textit{et al.}~\cite{kalofolias2016} proposed a convex formulation as follows
\begin{equation}
  \begin{array}{ll}
    \underset{\bm{W}}{\mathsf{minimize}}&
    \frac{1}{2}\tr\left({\bm{W}\bm{Z}}\right)
    -\alpha\mathbf{1}^\top\log(\bm{W}\mathbf{1}) +
    \frac{\gamma}{2}\norm{\bm{W}}^{2}_{\mathrm{F}}, \\
    \mathsf{subject~to} & W_{ij} = W_{ji} \geq 0, \mathsf{diag}(\bm{W}) = \mathbf{0},
  \end{array}
  \label{eq:smooth_static_graph}
\end{equation}
where $Z_{ij} \triangleq \norm{\bm{x}_i - \bm{x}_j}^2_2$.

Problem~\eqref{eq:smooth_static_graph} is convex and can be solved via primal-dual, ADMM-like
algorithms~\cite{kalofolias2016}. %~\cite{komodakis2015}.
It can be seen that the objective function in Problem~\eqref{eq:smooth_static_graph} is actually
an approximation to that of Problem~\eqref{eq:laplacian-learning} where the $\mathsf{gdet}(\bm{L})$ term has been
upper bounded by $\prod_{i=1}^{p}L_{ii}$. Therefore, Problem~\eqref{eq:smooth_static_graph}
can be thought of as an approximation of the penalized maximum likelihood estimator.

A formulation to estimate $k$-component graphs based on the smooth signal approach was proposed in~\cite{feiping2016}.
They proposed a two-stage algorithm where it first estimates a connected graph using, \textit{e.g.}, Problem~\eqref{eq:smooth_static_graph}
and then it projects the graph onto the set of Laplacian matrices of dimension $p$ with rank $p-k$, where $k$ is the given number of graph components.

Spectral constraints on the Laplacian matrix are an intuitive way to recover $k$-component graphs as the multiplicity
of its zero eigenvalue, \textit{i.e.}, the nullity of $\bm{L}$, dictates the number of components of a graph. 
The first proposed framework to impose structures on the estimated Laplacian matrix, under the attractive improper GMRF model,
was done by Kumar~\textit{et al.}~\cite{kumar20192, kumar2019}, through the use of spectral constraints, as follows
\begin{equation}
  \begin{array}{ll}
    \underset{\bm{L}, \bm{U}, \boldsymbol\lambda}{\mathsf{minimize}} & \mathsf{tr}\left(\bm{L}\bm{S}\right)
    - \displaystyle\sum_{i=1}^{p-k}\mathsf{log}\left(\lambda_{i}\right) + \frac{\eta}{2} \norm{\bm{L} - \bm{U}\mathsf{Diag}(\boldsymbol\lambda)\bm{U}^\top}^{2}_{\mathrm{F}},\\
    \mathsf{subject~to} & \bm{L} \succeq \mathbf{0},~\bm{L}\mathbf{1} = \mathbf{0},~L_{ij} = L_{ji} \leq 0,\\
    & \bm{U}^\top\bm{U} = \bm{I}, \bm{U} \in \mathbb{R}^{p\times (p-k)}, \\
    & 0 < c_1 < \lambda_{1} < \dots < \lambda_{p-k} < c_2.
  \end{array}
  \label{eq:gmrf}
\end{equation}

Note that Problem~\eqref{eq:gmrf} learns a $k$-component graph without the need for a two-stage algorithm.
However, a clear shortcoming of this formulation is that it does not control the degrees of the nodes in the graph,
which may result in a trivial solution that contains isolated nodes, turning out not to be useful for
clustering tasks especially when applied to noisy data sets.

\section{Graph Laplacian Constraints Interpretation for Stock Signals}
Graphical representations of data are increasingly important tools in financial signal processing %~\cite{deprado2016}
to uncover hidden relationships between variables~\cite{marti2017}. In stock markets, one is generally interested
in learning about conditional dependencies among stocks and how to leverage this information
into practical scenarios such as portfolio design and crisis forecasting.

Mathematically, we would like to estimate a precision matrix $\bm{L}$ that
enjoys the following two key properties
\begin{description}
  \item[(P1)] $\bm{L}\mathbf{1} = \mathbf{0}$,
  \item[(P2)] ${L}_{ij} = {L}_{ji} \leq 0 ~\forall ~i \neq j$.
\end{description}

The first property states that the Laplacian matrix $\bm{L}$ is singular and its null space
contains the $\mathbf{1}$ vector. That means that any signal sampled from $\bm{L}$, say $\bm{x}$,
is constrained to a subspace of rank $p-1$ satisfying $\bm{x}^\top \mathbf{1} = 0$. In practice, (P1)
implies that signals living in a graph $\mathcal{G}$ have zero graph-mean.

Property (P2) together with (P1) implies that $\bm{L}$ is positive semidefinite.
The fact that the off diagonal entries are non-positive means that the Laplacian matrix only represents non-negative conditional
dependencies\footnote{The correlation between any two pair of nodes conditioned on the rest of the graph is given as $-\frac{L_{ij}}{\sqrt{L_{ii}L_{jj}}}$.}.
This assumption is often met for stock data as assets are typically positively dependent~\cite{agrawal2019,wang2020}.

These two properties along with efficient learning frameworks make the Laplacian-based graphical model
a natural candidate for learning graphs of stock data. As a consequence of using the Laplacian model,
we propose the following guidelines when estimating Laplacian matrices with stock market data.

\begin{itemize}
\item \textbf{Correlation vs Covariance}: Both the GMRF and smooth signal approaches rely on the term
$\tr{(\bm{S}\bm{L})} \propto \tr(\bm{W}\bm{Z})$, where $\bm{S}$ is the sample covariance matrix.
From the definition of $\bm{Z}$~\eqref{eq:smooth_static_graph}, we observe that two perfectly correlated stocks but with large Euclidean distances would
appear very different on the graph.
Hence, we should use the
sample correlation matrix $\bar{\bm{S}} = \mathsf{Diag}(\bm{S})^{-1/2}\bm{S}\mathsf{Diag}(\bm{S})^{-1/2}$
in case we want two highly correlated stocks to have a strong graph connection regardless of their variances.
\item \textbf{Removing the market trend}: 
A widely used and tested model for the returns of the stocks includes explicitly the dependency on the market factor:
$\bm{x}_t = \boldsymbol{\beta}x_{\mathsf{mkt},t} + \boldsymbol\epsilon_t$,
where $\boldsymbol{\beta}$ denotes the market loadings, $x_{\mathsf{mkt},t}$ denotes the market index, and $\boldsymbol\epsilon_t$
is the residual idiosyncratic component with covariance
matrix $\boldsymbol{\Psi}$. Since all the stocks are heavily dominated by the market index $x_{\mathsf{mkt},t}$, it
may be convenient to remove that component if we seek to explore the structure of the residual cross-dependency among the
stocks, $\boldsymbol\epsilon_t$. Thus, an alternative to using the full covariance matrix $\boldsymbol{\Sigma}$ is to use the
covariance matrix $\boldsymbol{\Psi}$ of the idiosyncratic component.
However, since $\boldsymbol{\beta}\approx\mathbf{1}$, it turns out that the market factor is automatically removed in the
squared distance matrix $\bm{Z}$: \begin{equation}
Z_{ij} \triangleq \Vert\bm{x}^{(i)}-\bm{x}^{(j)}\Vert^2_2 \approx \Vert\boldsymbol\epsilon^{(i)}-\boldsymbol\epsilon^{(j)}\Vert^2_2.\end{equation}
Even more interestingly, if one first normalizes each stock, whose variances are $\mathbb{V}(\bm{x}^{(i)}) \approx \beta^2_{i}$,
we have $\bar{\bm{x}} = \mathbf{1}\bar{x}_{\mathsf{mkt}} + \bar{\boldsymbol{\epsilon}}_t$, which implies that the market factor is
automatically removed in the squared distance matrix.
\item \textbf{Degree control}: Enforcing a rank smaller than $p-1$ for the Laplacian matrix will generate a $k$-component graph,
which is one desired goal. However, one may get the undesired result of having isolated nodes. The way to avoid isolated
nodes is by controlling the degrees of the nodes. The GMRF formulation has the natural penalty term $\mathsf{log~gdet}(\bm{L})$
in the objective, but that does not help in controlling the degrees of the nodes. Instead, some of the graph learning
formulations from smooth signals include degree control via the constraint $\bm{W}\mathbf{1}=\mathbf{1}$,
which fixes the degrees of all the nodes to $1$. The regularization term $\mathbf{1}^\top \mathsf{log}(\bm{W}\mathbf{1})$
also avoids the trivial solution of any degree equals $0$.
Hence, any graph learning formulation that enforces a $k$-component graph
(or low-rank Laplacian matrix) should also control the degrees of the nodes to avoid a trivial solution with isolated nodes.
\end{itemize}

\section{Proposed Formulations}

In this section, we propose graph learning formulations to account for:
1) $k$-component structures
and 2) non-stationarity of financial stock market data.

\subsection{$k$-component graphs: GMRF formulation}

We propose the following formulation to learn a $k$-component graph:
\begin{equation}
  \begin{array}{ll}
    \underset{\bm{L} \succeq \mathbf{0}}{\mathsf{minimize}} & \mathsf{tr}\left(\bm{L}\bm{S}\right)
    - \mathsf{log~gdet}\left(\bm{L}\right) \\
    \mathsf{subject~to} & \bm{L}\mathbf{1} = \mathbf{0},~L_{ij} = L_{ji} \leq 0, ~\forall~ i\neq j, \\
                        & \mathsf{diag}(\bm{L}) = \mathbf{1}, \mathsf{rank}(\bm{L}) = p-k.
  \end{array}
  \label{eq:k-comp-proposed}
\end{equation}

Problem~\eqref{eq:k-comp-proposed} is highly non-convex and non-differentiable due to the
constraint
$\mathsf{rank}(\bm{L}) = p-k$, which is equivalently to saying that
the sum of the $k$ smallest eigenvalues of $\bm{L}$ is equal to zero, \textit{i.e.},
$\sum_{i=1}^{k}\lambda_{i}\left(\bm{L}\right) = 0$~\cite{feiping2016} (assuming eigenvalues in increasing order).
By Fan's theorem~\cite{fan1949},
\begin{equation}\sum_{i=1}^{k}\lambda_{i}\left(\bm{L}\right) =
\underset{\bm{V} \in \mathbb{R}^{p\times k}, \bm{V}^\top \bm{V} = \bm{I}}{\mathsf{minimize}} \mathsf{tr}\left(\bm{V}^\top \bm{L} \bm{V}\right).
\end{equation}

Thus, a relaxed version of Problem~\eqref{eq:k-comp-proposed} becomes
\begin{equation}
  \begin{array}{ll}
    \underset{\bm{L} \succeq \mathbf{0}, \bm{V} \in \mathbb{R}^{p\times k}}{\mathsf{minimize}} & \mathsf{tr}\left(\bm{L}\bm{S}\right)
    - \mathsf{log~gdet}\left(\bm{L}\right) + \eta\mathsf{tr}\left(\bm{V}^\top \bm{L} \bm{V}\right),\\
    \mathsf{subject~to} & \bm{L}\mathbf{1} = \mathbf{0},~L_{ij} = L_{ji} \leq 0, ~\forall~ i\neq j, \\
                        & \mathsf{diag}(\bm{L}) = \mathbf{1}, \bm{V}^\top\bm{V} = \bm{I}.
  \end{array}
  \label{eq:k-comp-proposed-relaxed}
\end{equation}

Even though still non-convex, but now differentiable, Problem~\eqref{eq:k-comp-proposed-relaxed} can be solved in an alternating fashion.
More precisely, for a given $\bm{L}$, say $\bm{L}^l$, we have the following subproblem for $\bm{V}$:
\begin{equation}
  \begin{array}{ll}
    \underset{\bm{V} \in \mathbb{R}^{p\times k}}{\mathsf{minimize}} &
    \mathsf{tr}\left(\bm{V}^\top \bm{L}^l \bm{V}\right),\\
    \mathsf{subject~to} & \bm{V}^\top\bm{V} = \bm{I},
  \end{array}
  \label{eq:k-comp-proposed-relaxed-sub-V}
\end{equation}
whose solution is given by the $k$ eigenvectors associated with the $k$ smallest eigenvalues of $\bm{L}^l$~\cite{absil2008}.

For a fixed value of $\bm{V}$, say $\bm{V}^l$, we have the following subproblem for $\bm{L}$:
\begin{equation}
  \begin{array}{ll}
    \underset{\bm{L} \succeq \mathbf{0}}{\mathsf{minimize}} & \mathsf{tr}\left(\bm{L}\left(\bm{S} + \eta \bm{V}^l {\bm{V}^l}^\top\right)\right)
    - \mathsf{log~gdet}\left(\bm{L}\right) \\
    \mathsf{subject~to} & \bm{L}\mathbf{1} = \mathbf{0}, L_{ij} = L_{ji} \leq 0, \mathsf{diag}(\bm{L}) = \mathbf{1}.
  \end{array}
  \label{eq:k-comp-proposed-relaxed-sub-L}
\end{equation}

Problem~\eqref{eq:k-comp-proposed-relaxed-sub-L} is convex and can be solved efficiently by, \textit{e.g.}, the methods developed
in~\cite{egilmez2017, zhao2019, kumar2019}.
Algorithm~\ref{alg:alg1} summarizes the implementation to solve Problem~\eqref{eq:k-comp-proposed-relaxed}.
\begin{algorithm}
  \caption{GMRF $k$-component graph learning}
    \label{alg:alg1}
    \SetAlgoLined
    \KwData{Similarity matrix $\bS$, initial estimate $\bm{L}^{0}$, rank constraint hyperparameter $\eta > 0$.}
    \KwResult{Laplacian estimation $\bm{L}$}
    \While{not converged}{
      update $\bm{V}^{l+1}$ by solving Problem~\eqref{eq:k-comp-proposed-relaxed-sub-V} fixing $\bm{L}$ at $\bm{L}^l$\\
      update $\bm{L}^{l+1}$ by solving Problem~\eqref{eq:k-comp-proposed-relaxed-sub-L} fixing $\bm{V}$ at $\bm{V}^{l+1}$\\
    }
\end{algorithm}

\subsection{Time-varying graphs}

Most graph learning frameworks are designed towards static graphs, which inherently neglect time-domain variations in real data.
As a result, they usually lack practicality for finance especially during nonstationary regimes, \textit{e.g.}, an economic crisis.

A time-varying graph is defined as a sequence of graphs stacked over time, \textit{i.e.},
$\left\{\mathcal{G}_{t}\right\}_{t=1}^{T} = \left\{\mathcal{V}_{t},
\mathcal{E}_{t}, \bm{W}_{t}\right\}_{t=1}^{T}$. We assume
the node set of each graph to be the same, \textit{i.e.},
$\mathcal{V}_{t} = \left\{1, 2, \dots, p\right\}, \forall ~t=1,2,\dots,T$.

The data generating process is as follows. Assume that for every graph $\mathcal{G}_{t}$
we associate an attractive improper GMRF $\bm{x}_{t} \sim \mathcal{N}\left(\mathbf{0}, \bm{L}^{\dagger}_{t}\right)$, where
$\bm{L}_{t}$ is the precision matrix of the $t$-th GMRF that is assumed to have graph Laplacian structure.
Further, suppose we are given $n_t$ observations from $\bm{x}_{t}$, \textit{i.e.}, $\bX_{t} \in \mathbb{R}^{n_t \times p}$, then we propose
the following optimization program to learn the Laplacian matrices $\left\{\bm{L}_{t}\right\}_{t=1}^{T}$
on the basis of $\left\{\bm{X}_{t}\right\}^{T}_{t=1}$, in particular $\bm{L}_t$ is obtained from
\begin{equation}
\begin{array}{ll}
  \underset{\bm{L}_{1}, \bm{L}_{2}, \dots, \bm{L}_{t} }{\mathsf{minimize}} &
  \sum_{\tilde{t}=1}^{t}
    n_{\tilde{t}}\left[\mathsf{tr}\left(\bm{S}_{\tilde{t}}\bm{L}_{\tilde{t}}\right)
    - \mathsf{log~gdet}\left(\bm{L}_{\tilde{t}}\right) \right] \\
  & + \delta \sum_{\tilde{t}=2}^{t}d\left(\bm{L}_{\tilde{t}}, \bm{L}_{\tilde{t}-1}\right),\\
  \mathsf{subject~to} & \left\{\bm{L}_{\tilde{t}} \succeq \mathbf{0}, \bm{L}_{\tilde{t}}\mathbf{1} = \mathbf{0}, (\bm{L}_{\tilde{t}})_{ij} = (\bm{L}_{\tilde{t}})_{ji} \leq 0\right\}_{\tilde{t}=1}^{t},
\end{array}
\label{eq:tv-graph}
\end{equation}
where $\{\bm{S}_{t}\}_{t=1}^{T}$ is a sequence of similarity matrices,
$d : \mathbb{R}^{p\times p}\times\mathbb{R}^{p\times p} \rightarrow \mathbb{R}_{+}$ is
a distance function that measures the similarity between
$\bm{L}_{t}$ and $\bm{L}_{t-1}$ in order to impose time consistency, \textit{e.g.},
$d(\bm{L}_{t}, \bm{L}_{t-1}) \triangleq \Vert \bm{L}_{t} - \bm{L}_{t-1} \Vert^{2}_{\text{F}}$, and
$\delta \in \mathbb{R}_{++}$ is its corresponding hyperparameter.
The solution to Problem~\eqref{eq:tv-graph} is obtained on a rolling window basis,
which is tantamount to a causal estimator $\hat{\bm{L}}_{t|t}, t=1,\dots,T$. In other words, to estimate, \textit{e.g.}, $\bm{L}_3$, we only use
information up to and including time $t=3$, \textit{i.e.}, $\{\bm{S}_{t}\}_{t=1}^{3}$.

The time-varying graph learning formulations proposed in~\cite{kalofolias2017} and~\cite{yamada2020} %yamada2019,}
are not adequate to our particular scenario because those formulations are based on the smooth signal approach,
rather than the statistical GMRF model, and solve a dynamic graph conditioned on all the $T$ chunks of observations, \textit{i.e.},
$\bm{\hat{L}}_{t \mid T}, t=1,\dots,T$, which inevitably introduces look-ahead biases.

Algorithm~\ref{alg:tv} summarizes the implementation to solve Problem~\eqref{eq:tv-graph}.
\begin{algorithm}
  \caption{Time-varying graph learning}
    \label{alg:tv}
    \SetAlgoLined
    \KwData{$\bm{S}_{1}, \dots, \bm{S}_{T}$, $\{n_t\}^{T}_{t=1}$}
    \KwResult{Causal Graph Laplacians estimates $\{\bm{\hat{L}}_{t\mid t}\}^{T}_{t=1}$}
    \For{$t=1:T$}{
      $\triangleright$ estimate $\{\bm{L}_{\tilde{t}}\}_{\tilde{t}=1}^{t}$ using data $\{\bm{S}_{\tilde{t}}\}^{t}_{\tilde{t}=1}$. \\
      $\triangleright$ store $\bm{L}_{t}$ as $\bm{\hat{L}}_t$. \\
    }
    \Return $\{\bm{\hat{L}}\}_{t=1}^{T}$.
\end{algorithm}

\noindent \textit{Remark}: for practical programming language implementation of the proposed algorithms,
one can take into account that the Laplacian matrix $\bm{L}$ is symmetric and its diagonal
elements are mappings of the off-diagonal ones. Therefore, only elements above (or below)
the diagonal of $\bm{L}$ need to be updated. This approach is similar to that used by
Kalofolias~\textit{et al.}~\cite{kalofolias2016} and Kumar~\textit{et al.}~\cite{kumar2019}.

\section{Numerical Experiments}

In the experiments that follow, we use log-returns time-series data from stocks belonging to three sectors of the
S\&P500 index, namely $\mathsf{Industrials}$, $\mathsf{Consumer~Staples}$, and $\mathsf{Energy}$. We collect
price data of 130 stocks from Jan. 1st 2016 to Jan. 1st 2019 from Yahoo! Finance, which represents
753 days worth of data.

\subsection{Effects of market factor and data scaling}

In order to measure the effects of the market factor and the data scaling, we learn four graphs using
the two-stage algorithm proposed by~\cite{feiping2016}.

\begin{figure}[!htb]
  \captionsetup[subfigure]{justification=centering}
  \centering
  \begin{subfigure}[t]{0.22\textwidth}
      \centering
      \includegraphics[scale=.33]{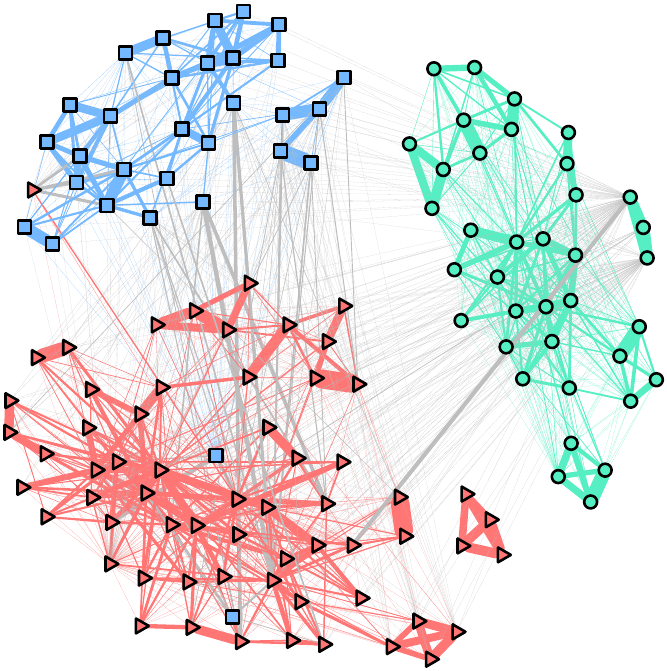}
      \caption{Market removed and no data scaling.}
  \end{subfigure}%
  ~
  \begin{subfigure}[t]{0.22\textwidth}
      \centering
      \includegraphics[scale=.33]{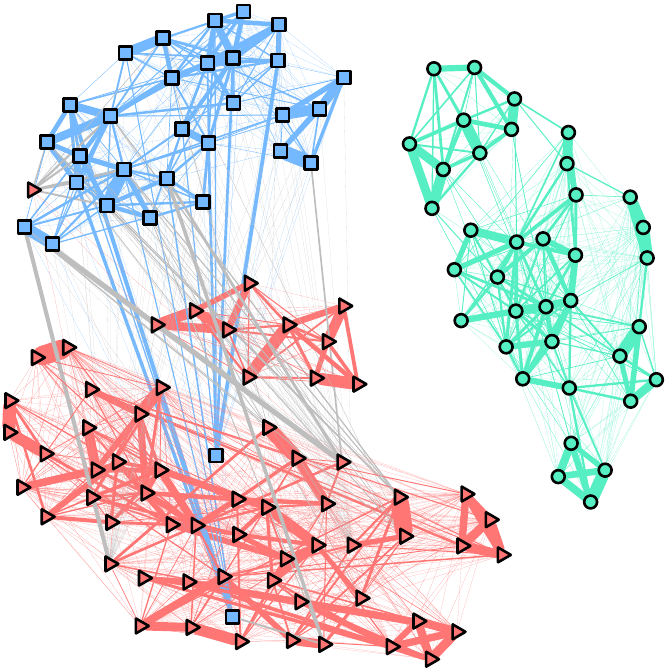}
      \caption{Market removed and data scaled.}
  \end{subfigure}%
  \\
  \begin{subfigure}[t]{0.22\textwidth}
      \centering
      \includegraphics[scale=.33]{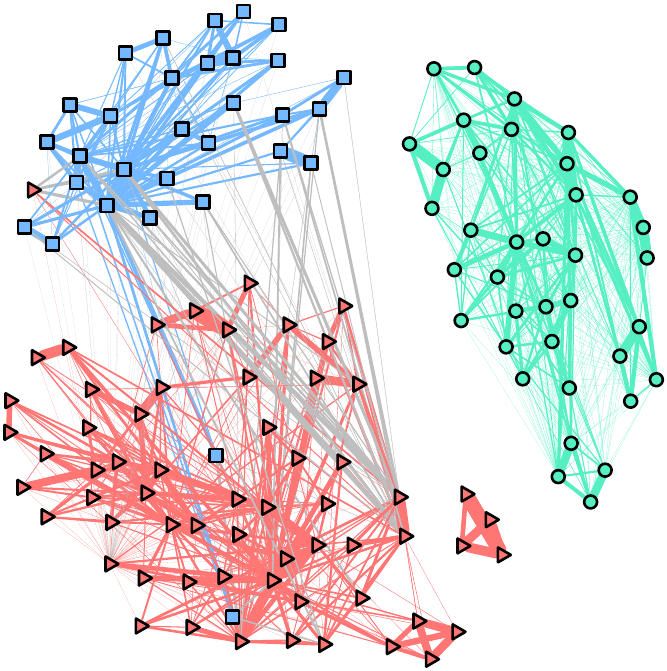}
      \caption{Market not removed and no data scaling.}
  \end{subfigure}%
  ~
  \begin{subfigure}[t]{0.22\textwidth}
    \centering
    \includegraphics[scale=.33]{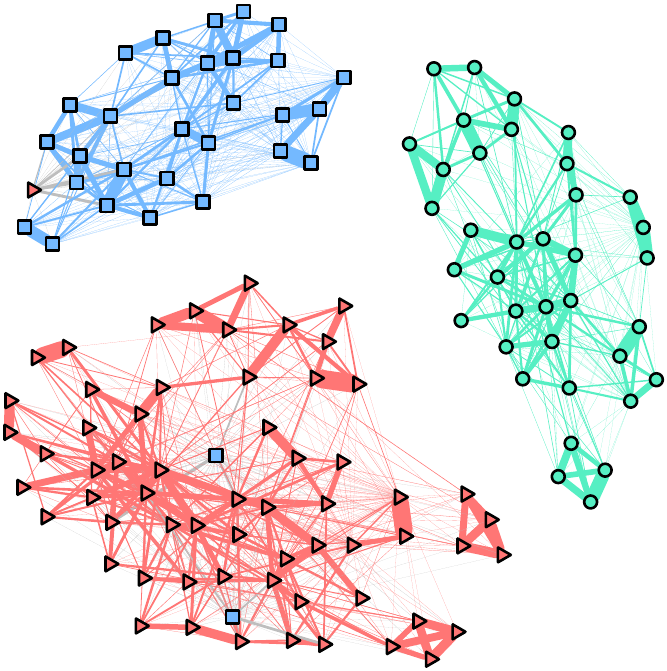}
    \caption{Market not removed and data scaling.}
\end{subfigure}
  \caption{Graphs estimated with different data preprocessing. Panel (d) gives the best results with scaled data and no market removed.}
  \label{fig:effects}
\end{figure}

Fig.~\ref{fig:effects}a shows that removing the market and using
the covariance matrix as input leads to a graph with many possibly spurious connections (grey edges). Fig.~\ref{fig:effects}b
shows that using the
sample correlation matrix introduces an improvement, but there exist still many spurious dependencies. Fig.~\ref{fig:effects}c
shows that
not removing the market presents an improvement, but the use of the covariance matrix leads to many possibly fake connections.
Fig.~\ref{fig:effects}d combines the correlation matrix as input and not removing the market, which clearly shows a meaningful representation
of a graph from stocks belonging to three distinct sectors.

\subsection{Effect of degree control}

\begin{figure}[!htb]
  \captionsetup[subfigure]{justification=centering}
  \centering
  \begin{subfigure}[t]{0.22\textwidth}
      \centering
      \includegraphics[scale=.4]{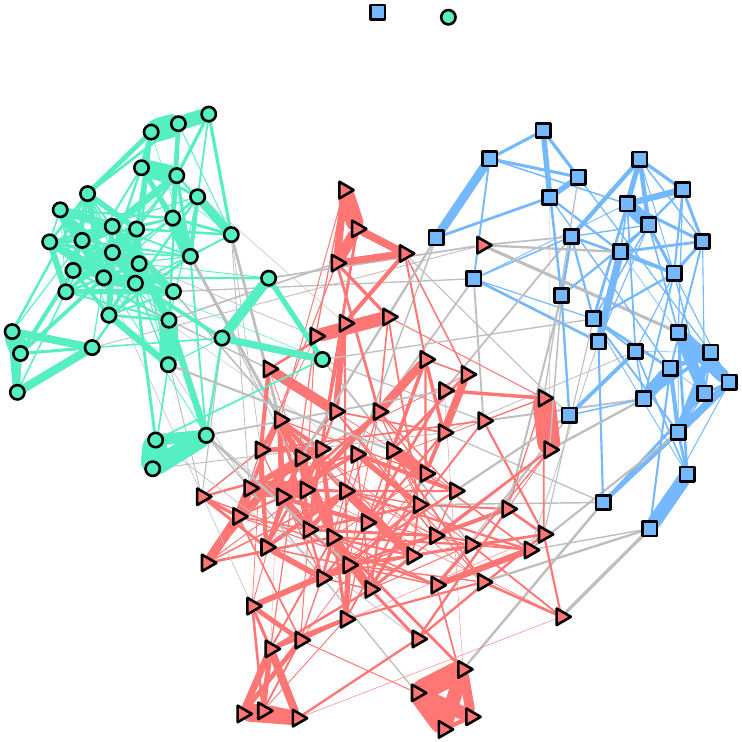}
      \caption{$\mathsf{SGL}$~\cite{kumar2019}.}
  \end{subfigure}%
  ~
  \begin{subfigure}[t]{0.22\textwidth}
    \centering
    \includegraphics[scale=.4]{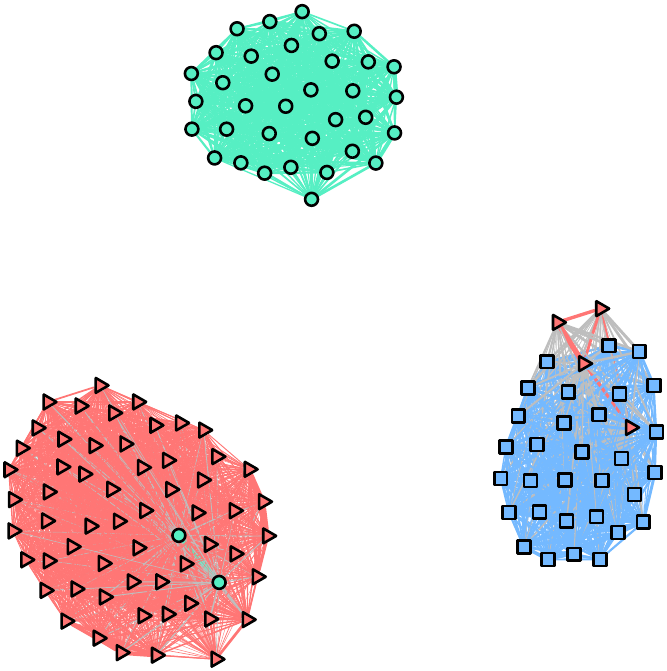}
    \caption{Algorithm~\ref{alg:alg1} (proposed).}
\end{subfigure}
  \caption{Effect of lack of degree control on the optimization formulation. Benchmark algorithm $\mathsf{SGL}$, panel (a),
           fails to obtain a meaningful solution, showing that spectral constraints alone are not sufficient to recover a non-trivial
           $k$-component graph. The proposed algorithm in panel (b) shows a meaningful estimated structure.
           We fixed the $\beta = 10$ for both methods.}
  \label{fig:sgl-failed}
\end{figure}

To illustrate the importance of controlling the nodes degrees while learning $k$-component graphs, we conduct
a comparison between the algorithm proposed in~\cite{kumar2019} and Algorithm~\ref{alg:alg1} on the basis
of the sample correlation matrix. Fig.~\ref{fig:sgl-failed} shows the estimated financial stock networks with
$k=3$. It is clear that $\mathsf{SGL}$~\cite{kumar2019}
(benchmark) is unable to account for the trivial solution with isolated nodes, whereas the proposed algorithm
returns a graph with a meaningful representation.

\subsection{Time-varying Experiment}
We consider an experiment with data from FAAMUNG companies
(Facebook, Apple, Amazon, Microsoft, Uber, Netflix, and Google) from June 1st 2019 to May 1st 2020, totalling 230 days worth of data,
which includes the most recent economic crisis due to the pandemic associated with COVID-19.
For each period of 30 days we estimate graphs on a rolling window fashion, shifting the window one day at a time,
on the basis of the sample correlation matrices $\bm{\bar{S}}_1, \dots, \bm{\bar{S}}_{200}$.
At the end, we estimate 200 graphs, \textit{i.e.}, $\bm{\hat{L}}_1, \dots, \bm{\hat{L}}_{200}$.

Based on this 200 estimated graphs we compute the algebraic connectivity, \textit{i.e.}, the second smallest eigenvalue of $\bm{L}_t$,
as an indicator %s
of variation of the graph. We use this indicator to acquire insights on possible trends
of the stock market.
Note that other indicators could be used in practice such as the spectral radius,
$\boldsymbol{\lambda}_{\mathsf{max}}(\bm{\hat{L}}_t)$,
and the time consistency, $\Vert \bm{\hat{L}}_t - \bm{\hat{L}}_{t-1} \Vert^{2}_{\text{F}}$.

\begin{figure}[!htb]
  \captionsetup[subfigure]{justification=centering}
  \centering
  \begin{subfigure}[t]{0.5\textwidth}
      \centering
      \includegraphics[scale=.385]{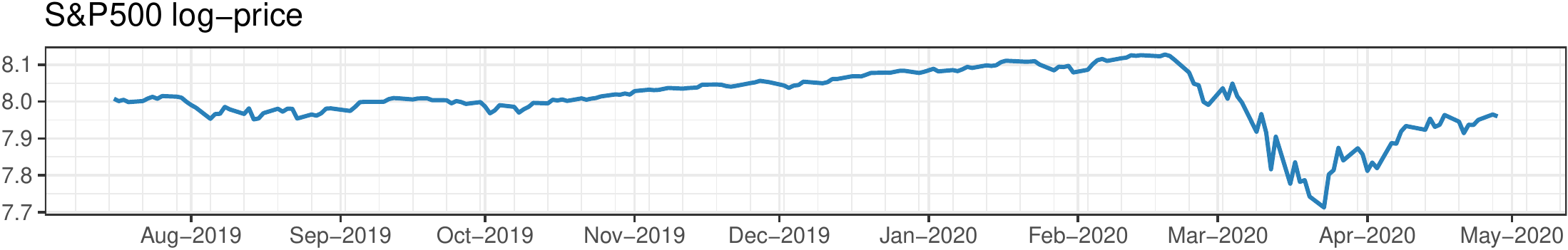}
      \caption{}
  \end{subfigure}%
  \\
  \begin{subfigure}[t]{0.5\textwidth}
    \centering
    \includegraphics[scale=.385]{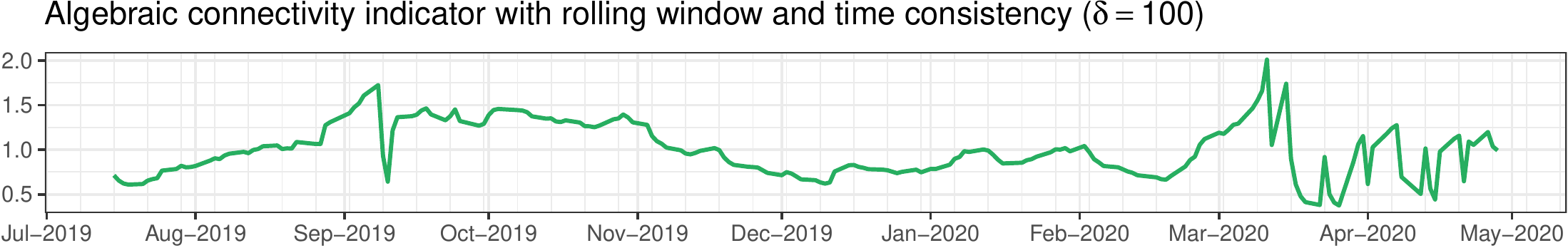}
    \caption{}
  \end{subfigure}%
  \\
  \begin{subfigure}[t]{0.5\textwidth}
    \centering
    \includegraphics[scale=.25]{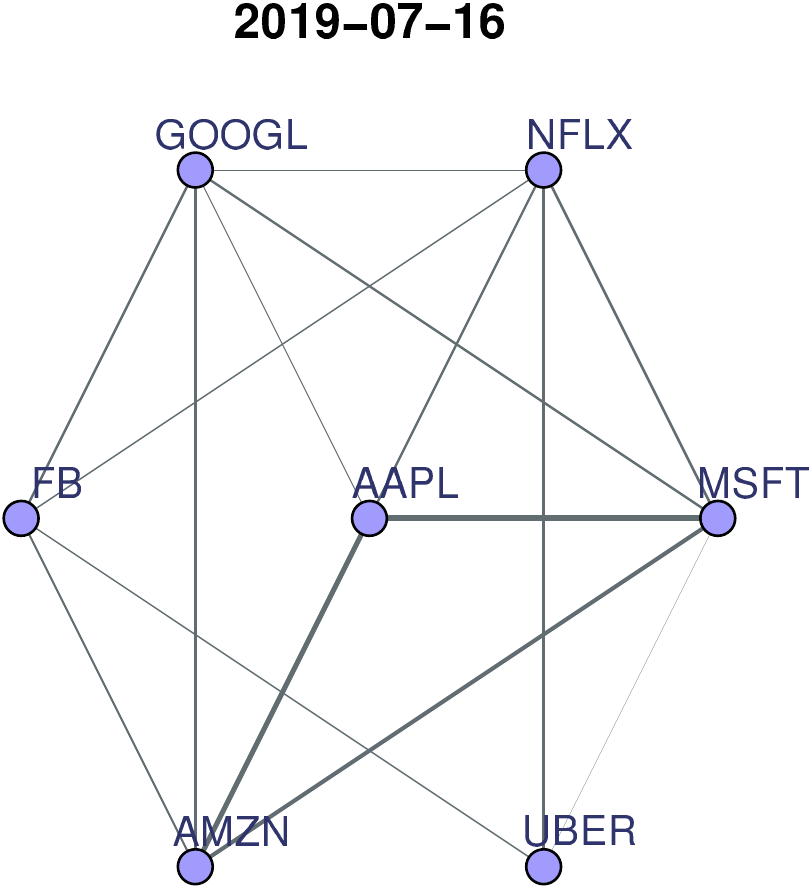}
    \includegraphics[scale=.25]{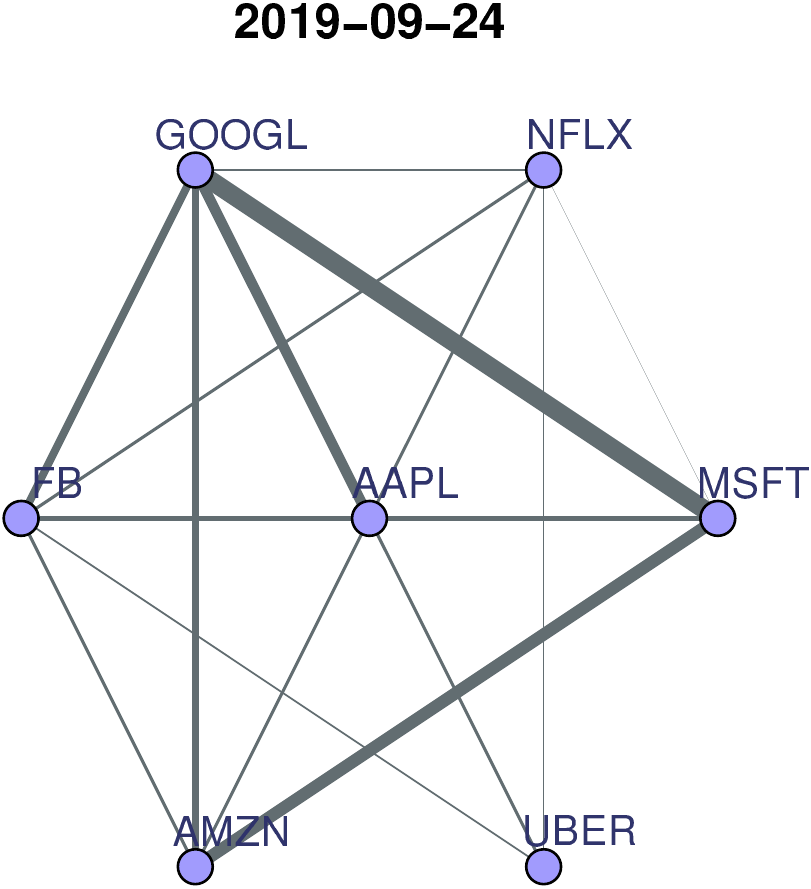}
    \includegraphics[scale=.25]{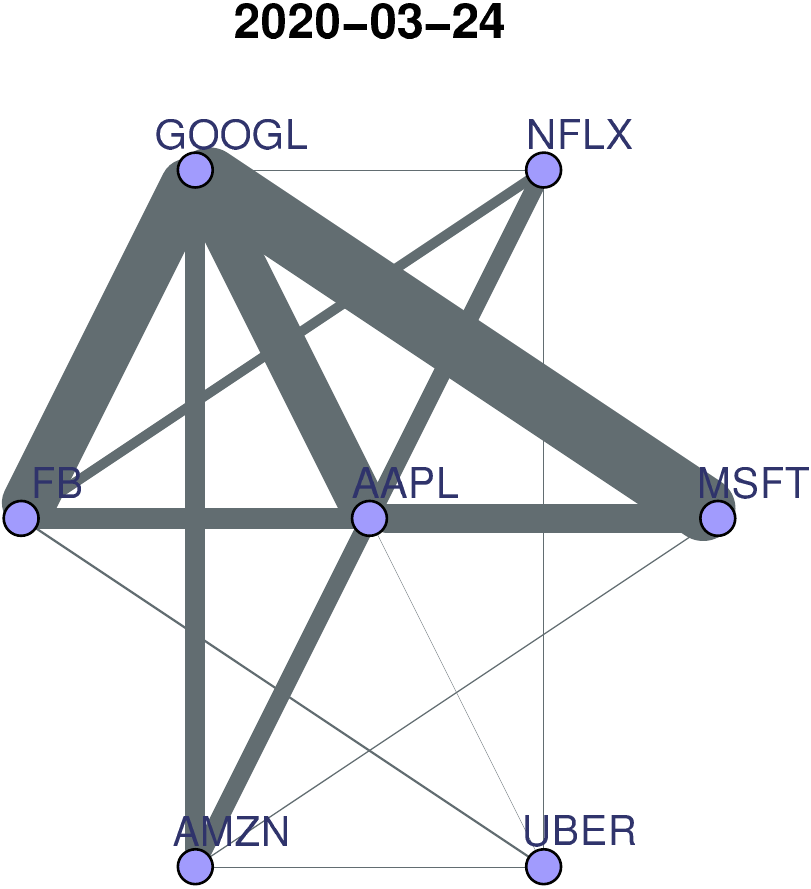}
    \includegraphics[scale=.25]{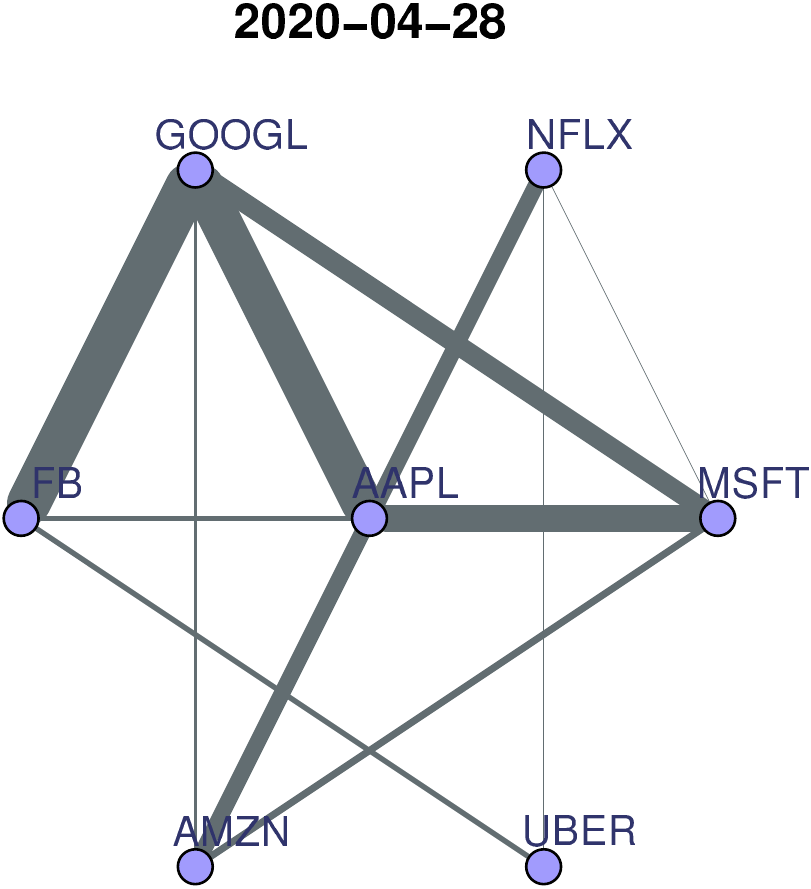}
    \caption{}
  \end{subfigure}
  \caption{Estimated time-varying indicators and network visualization of FAAMUNG companies during 2019-2020.
  Panel (a) shows the log-price of the S\&P500 index.
  Panel (b) show the algebraic connectivity indicator
  for $\delta = 100$. Panel (c) shows the estimated networks with $\delta = 100$
  for several dates. It can be noticed that there is an increase in conditional correlation among the stocks during
  times of economic crisis.}
  \label{fig:tv-indicators}
\end{figure}

Fig.~\ref{fig:tv-indicators}a shows the S\&P500 log-price, where the impact of the COVID-19 pandemic
is clear around March 2020.
Fig.~\ref{fig:tv-indicators}b depicts the algebraic connectivity indicator computed from each estimated graph along the time axis.
Fig.~\ref{fig:tv-indicators}c shows the evolution of the graph network at certain dates. It is clear both from the indicator
(Fig.~\ref{fig:tv-indicators}b)
and the network visualization (Fig.~\ref{fig:tv-indicators}c)
that around September 2019 the market has changed significantly. That is consistent
with news involving the impeachment inquiry of US President Donald J. Trump. From the middle of March 2020
to the beginning of May the market saw its largest drop since the financial crisis in 2008. This can also
be noticed through the indicator and the network visualization.

\subsection{Trading Application}
\begin{figure}[!htb]
  \centering
  \includegraphics[scale=.385]{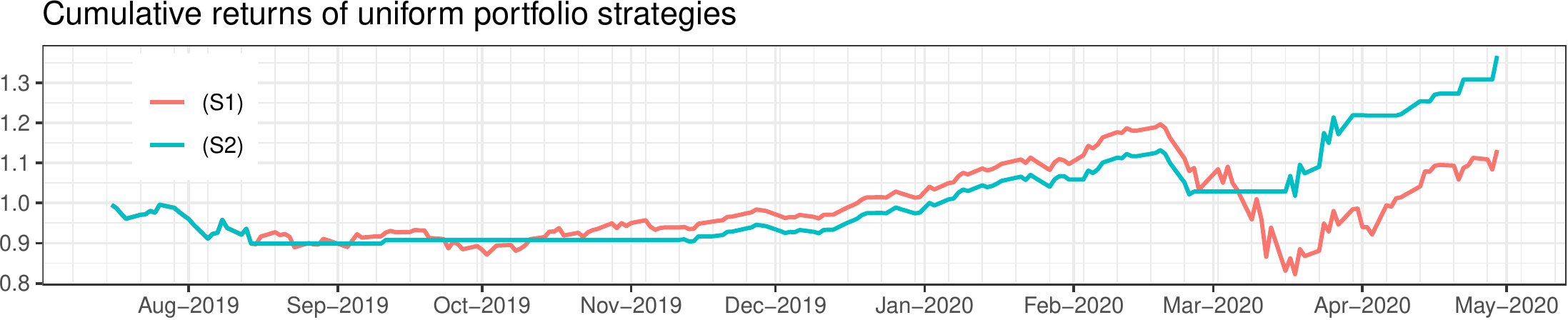}
  \caption{Uniform investment strategies on FAAMUNG companies. The cumulative PnL of strategy (S2),
  which enters/exits the market according to the knowledge of algebraic connectivity of the estimated graph,
  outperforms that of strategy (S1), which continuously invests over the whole time period.
  }
  \label{fig:portfolio}
\end{figure}

By leveraging the proposed time-varying graph learning algorithm, we perform an experiment comparing two simple trading
strategies: (S1) uniformly invest a unit of budget during the whole period;
(S2) uniformly invest a unit of budget according to whether or not the algebraic connectivity of the graph falls below a fixed
threshold of $\tau = 1.0$.
Fig.~\ref{fig:portfolio} shows the cumulative sum of the profits and losses (PnL) over time for (S1) and (S2).
It can be observed that (S2) outperforms (S1) by smartly entering/exiting the market based on the algebraic connectivity
of the estimated graphs.

%We would like to thank Pedro for the BBQs.

\bibliographystyle{IEEEtran}
\bibliography{vini-graph-asilomar}

\end{document}